# Comparative Analysis of the Land Use and Land Cover Changes in Different Governorates of Oman using Spatiotemporal Multi-spectral Satellite Data

Muhammad Shafi [1,3], Syed Mohsin Bokhari [2]

[1] School of Computing, Ulster University, Belfast, UK; m.shafi@ulster.ac.uk
[2] University of Engineering and Technology Taxila; smohsin599@gmail.com
[3] Faculty of Computing and Information Technology, Sohar University
\* Correspondence: mfnoor@gmail.com

**Abstract:** Land cover and land use (LULC) changes are key applications of satellite imagery, and they have critical roles in resource management, urbanization, protection of soils and the environment, and enhancing sustainable development. The literature has heavily utilized multispectral spatiotemporal satellite data alongside advanced machine learning algorithms to monitor and predict LULC changes. This study analyzes and compares LULC changes across various governorates (provinces) of the Sultanate of Oman from 2016 to 2021 using annual time steps. For the chosen region, multispectral spatiotemporal data were acquired from the open-source Sentinel-2 satellite dataset. Supervised machine learning algorithms were used to train and classify different land covers, such as water bodies, crops, urban, etc. The constructed model was subsequently applied within the study region, allowing for an effective comparative evaluation of LULC changes within the given timeframe.

## 1. Introduction

One of Oman's most important objectives is the diversification of its economic base and the reduction of dependence on oil. Contemporary urban planning and the development of smart cities are critical in supporting economic diversification and sustainable development. In this regard, this present study also seeks to improve future urban planning and maximize the management of resources in the country.

Although land use and land cover are often regarded as synonymous terms, they are actually different. Land use refers to human uses of an area, which is not always apparent from satellite images. Land cover, on the other hand, is the physical and biological features on the Earth's surface, which can be identified from remote sensing approaches. Information on LULCC is critical to examine past developments and forecast land use trends for the future. Such information is relevant in most areas of resource management, environmental sustainability, urban planning, agricultural development, and smart city development. Studies on LULCC primarily answer fundamental questions such as Where and why are changes in land cover occurring? What are the nature and extent of these changes? What are the forces behind these changes? Moreover, what are trends expected for land cover changes in the future? A number of studies have employed satellite imagery in the study of land use changes and land cover changes (LULCC) in different geographical settings. The following text provides a brief review of relevant literature, taking into consideration methodologies, data sources, analysis methods, and significant conclusions in LULCC analysis using satellite imagery.

Most studies in this field utilize satellite and remote sensing data, and some of them use spatiotemporal data obtained from in-situ techniques, e.g., drone images. There are numerous satellite and remote sensing datasets, with a distinction between open access and subscription datasets. Table 1 presents some commonly used satellite datasets applicable to land use and land cover change (LULCC) analysis.

| Satellite | Spatial Resolution | Temporal Resolution | Coverage | Bands | Free Availability |
|---|---|---|---|---|---|
| Sentinel-2 | 10m, 20m, 60m | 5 days | Global | 13 (incl. RGB, NIR, SWIR) | Yes |
| Landsat 8/9 | 30m (15m Pan) | 16 days | Global | 11 (incl. RGB, NIR, SWIR, Thermal) | Yes |
| MODIS | 250m, 500m, 1km | 1-2 days | Global | 36 (Visible to Thermal IR) | Yes |
| GeoEye-1 | 0.41m (Pan), 1.65m (Multi) | 3 days | On-demand | 5 (Pan + RGB + NIR) | No |
| WorldView-3 | 0.31m (Pan), 1.24m (Multi) | <1 day | On-demand | 8 (Pan + RGB + NIR + SWIR) | No |
| SPOT 6/7 | 1.5m (Pan), 6m (Multi) | 1 day | On-demand | 5 (Pan + RGB + NIR) | No |
| Sentinel-1 | 5m - 40m | 6 days | Global | 2 (C-band SAR) | Yes |
| Planet SkySat | 0.5m (Pan), 1m (Multi) | Daily | On-demand | 5 (Pan + RGB + NIR) | No |
| ASTER | 15m, 30m, 90m | 16 days | Global | 14 (Visible to Thermal IR) | Yes |
| Sentinel-3 | 300m - 1km | <2 days | Global | 21 (Visible to Thermal IR) | Yes |

**Table 1.** Satellite characteristics including spatial and temporal resolution, coverage, bands, and free availability.

The Table 1 offers a comparative evaluation of various Earth observation satellites, including their spatial and temporal resolutions, coverage, spectral bands, and availability of data for free. Sentinel-2, Landsat 8/9, MODIS, ASTER, and Sentinel-3 offer global coverage with freely available data and are therefore significant for environmental monitoring and research. Specifically, Sentinel-2 produces high-resolution images at 10m, 20m, and 60m at a revisit frequency of five days, whereas Landsat 8/9 offers a resolution of 30m (15m panchromatic) and works at a 16-day revisit cycle. On the other hand, MODIS, ASTER, and Sentinel-3 produce lower resolution images but offer frequent revisits and cover a wide spectral range including thermal infrared. On the other hand, commercial satellites like GeoEye-1, WorldView-3, SPOT 6/7, and Planet SkySat offer very high spatial resolutions of up to 0.31m (panchromatic). However, their data are available only through on-demand requests and are therefore limited from open research applications. Moreover, Sentinel-1, a synthetic aperture radar (SAR) satellite, offers all-weather imagery with resolutions of between 5m and 40m.

Satellite data has been extensively used in the analysis of land use and land cover change (LULCC) due to its high spatial coverage, temporal availability, and spectral richness. Numerous remote sensing techniques, including pixel-based as well as object-based classification procedures, have been used by researchers to extract land cover information from satellite data. For example, [1] used pixel-based classification to Landsat data to estimate LULCC in the Pearl River Delta, China, to reveal widespread urban expansion. Likewise, object-based classification procedures have been used to integrate spatial data and enhance the accuracy of classification [2]. [3] used high-spatial-resolution Landsat data along with object-based classification to analyze LULCC in a coastal region of Kerala, India. Change detection techniques are widely employed to detect and quantify LULCC from satellite data. Post-classification comparison techniques compare land cover maps at different temporal periods to determine changes. For example, [4] used post-classification comparison to analyze LULCC in the Three Gorges Reservoir of China to map significant land cover changes, such as urban expansion and land use conversion to agriculture. In addition, image differencing techniques have been employed to detect land cover attribute changes [5]. In a study [6], image differencing was used to monitor the conversion of wetlands to urban land in the Pearl River Delta, China. Spectral indices, including the Normalized Difference Vegetation Index (NDVI), are frequently utilized to track vegetation change and detect land cover transitions [7]. For instance, [8] utilized NDVI to examine land cover changes in China's Yangtze River Delta, which showed widespread conver-- sions of croplands to built-up land. LULCC research utilizing satellite imagery has been conducted in various geographical regions and ecosystems. Large-scale evaluations have documented large-scale land cover changes linked to urbanization, deforestation, and agricultural expansion [9]. In [10] examined land use change dynamics in China's rapidly developing Tianjin Binhai region, which showed widespread urban expansion and land use change.

Monitoring deforestation and forest degradation is one of the important applications of satellite imagery, which detects hotspots of deforestation, drivers, and environmental impacts [11]. For instance, [12] estimated forest cover loss in Brazil's Amazon using satellite imagery, which showed drivers of infrastructure development and land clearing. Monitoring agricultural land and crops has also utilized satellite imagery to evaluate changes in agricultural practices, productivity, and food security [13]. In [14] used satellite data to map crop types and rotations in the Argentine Pampas to support enhanced agricultural planning and sustainability. Satellite-based research has also played an important role in monitoring ecosystem dynamics, biodiversity loss, and conservation planning initiatives [15].

A sizeable number of studies have discussed the use of Geographic Information Systems (GIS) and spatiotemporal data analysis of various land-use studies in Oman. In [16] analyzed five mountain oases through GIS-based land surveys and Google Earth information and concluded that the areas under crops have drastically reduced due to drought and irrigation issues. A comparable study was carried out by Luedeling and Buerkert [17] in the Jabal Al Akhdar area, where supervised classification of Landsat and Ikonos imagery was applied to evaluate land use and land cover change. [18] gathered geospatial data on greenhouses and farms based on GPS-based georeferencing in nine Wilayats of Oman. Their results indicated that intensive agriculture is moving away from quickly urbanizing areas and high groundwater salinity areas, resulting in the development of large inland farms and abandonment of traditional coastal farms. Land use changes in different provinces, such as Braka, Wilayat Nizwa, Musandam, and Dhofar, have been examined by Caspari [19], Donato, and Jendryke (2019); Fadda, Al Shebli, and Al Kabi [20]; Megdiche-Kharrat et al. [21]; and Ramadan, Al-Awadhi, and Charabi [22]. The effects of extreme weather events on land use patterns in Oman have also been examined. Al-Hatrushi and Al-Alawi [23] evaluated the effects of two severe tropical cyclones—Gonu (June 2007) and Phet (June 2010)—that had a powerful impact on land use in Oman. According to their results, the recreational and residential sectors were the most impacted, each contributing 31% of the overall impact. Moderate impacts were found on public buildings, agricultural lands, and road facilities.

In spite of the increasing volume of research on LULCC, research on urbanization in Oman is scarce. Bridging this gap, the current study tries to examine the spatiotemporal dynamics of land use change in Oman by integrating satellite data and sophisticated GIS algorithms. It is essential to understand the spatiotemporal dynamics of LULCC and forecast future urbanization patterns in order to facilitate sustainable development, make national policy, and create efficient urban planning strategies.

The primary objectives of this study are to:

- Extract and preprocess relevant satellite imagery for the study area within the specified timeframe.
- Delineate and crop the study area for detailed analysis.
- Implement supervised machine learning classification to categorize and identify urban land cover.
- Analyze the results to derive insights into urban expansion and predict potential future impacts.

## 2. Materials and Methods

Figure 2 provides the flowchart of the research methodology. Different aspects of the methodology such as the study area, used satellite data, land cover labels, and accuracy assessment are discussed below.

*2.1. Study Area*

Oman is an Arabian state bordering Saudi Arabia, Yemen, the United Arab Emirates by land, and Pakistan and Iran by sea. Oman is 309,500 km² in area and is inhabited by over 4.5 million people. Oman is situated in the southeastern region of the Arabian Peninsula, with geographical coordinates between 16.40° and 26.20° north latitude and 51.50° and 59.40° east longitude. The economy of Oman is greatly dependent on oil exports, as it is one of the world's top oil producers. Oman is further divided into 11 administrative regions or governorates: Muscat, Dhofar, Musandam, Buraymi, the Dakhiliyah, North Batinah, South Batinah, South Sharqiyah, North Sharqiyah, Dhahirah, and Wusta. The largest governorate in the area is Dhofar, and the most populated is Muscat. Table 2 presents comprehensive information regarding the various governorates, including location and population as shown in Figure 1.

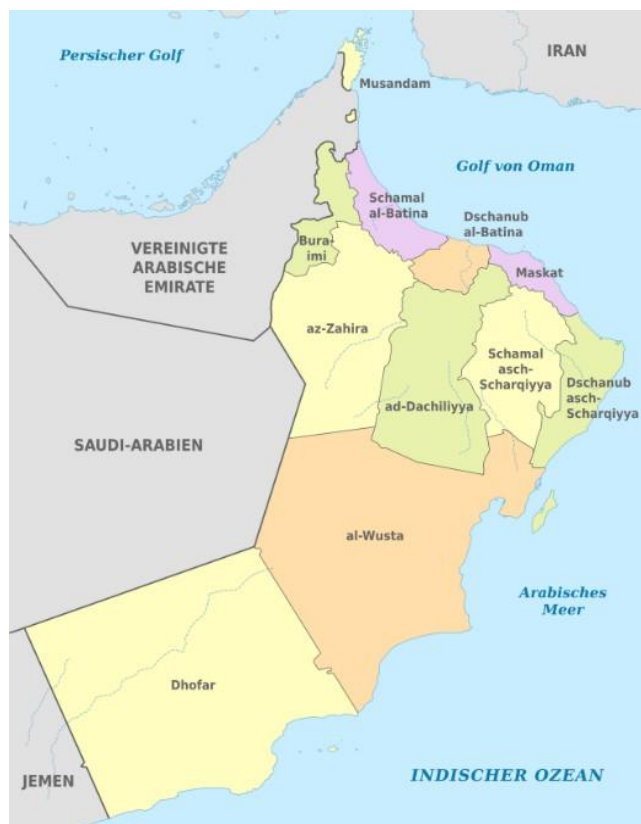

**Figure 1.** The Geographic Location of Oman.

*2.2. Satellite Data Description*

The satellite data used in this research belongs to the years 2017, 2018, 2019, 2020, and 2021. The source data are in the Universal Transverse Mercator (UTM) WGS84 coordinate reference system, and the service coordinate reference system is Web Mercator Auxiliary Sphere WGS84 (EPSG:3857) compatible. The geographical extent of the data is the entire area of Oman. Sentinel-2 L2A imagery was used as the base imagery to classify land cover with a cell size of 10 meters, which provides a fine and high-resolution depiction of the ground. The thematic attributes of the data are compatible with the classification of different land cover classes based on the spectral features sensed by the satellite.

| Governorate | Capital | Population (2020) | Area (km²) | Number of Wilayats |
|---|---|---|---|---|
| Muscat | Muscat | 1,302,440 | 3,500 | 6 |
| Al Batinah North | Sohar | 1,250,231 | 9,000 | 6 |
| Dhofar | Salalah | 416,458 | 99,300 | 10 |
| Ad Dakhiliyah | Nizwa | 478,501 | 31,900 | 8 |
| Ash Sharqiyah South | Sur | 315,445 | 12,039 | 5 |
| Ash Sharqiyah North | Ibra | 271,822 | 24,361 | 6 |
| Ad Dhahirah | Ibri | 213,043 | 37,000 | 3 |
| Al Buraimi | Al Buraimi | 121,802 | 7,460 | 3 |
| Al Wusta | Haima | 52,344 | 79,700 | 4 |
| Musandam | Khasab | 49,062 | 1,800 | 4 |

**Table 2.** Governorates of Oman with their capitals, population, area, and number of Wilayats.

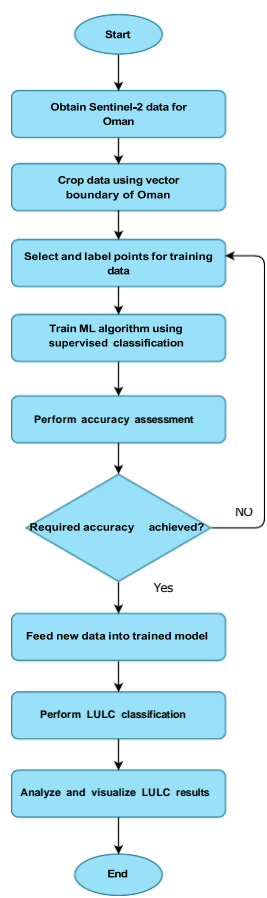

**Figure 2.** Flowchart of the Research Methodology.

*2.3. The Land Cover Labels*

The types of land covers studied in this research are grouped into several distinct types. Water defines lands that are predominantly characterized by water for most of the year, with little vegetation and no man-made features. Examples include aquatic land features such as rivers, lakes, ponds, and oceans. Tree defines lands dominated by high and dense arboreal cover (usually more than 15 feet), such as forests, wooded vegetation, dense clusters of vegetation in savannas, plantations, wetlands, and mangroves. Crops define agricultural lands cultivated by humans, which consist mainly of crops such as maize, wheat, sugarcane, and tobacco. These lands are usually without high tree height and have well-organized land use patterns suitable for agriculture. The Built-up Area class defines anthropogenic constructions, which comprise residential structures, roadways, buildings, railway networks, towns, and cities. Finally, Bare Grounds define lands without vegetation for the entire year, such as rocky ground, sandy deserts, bare ground, and lands covered by mining or dried-up lakes. These lands are usually characterized by little to no vegetation and are dominated by geological features such as rocks and sand.

*2.4. Accuracy Assessment*

Accuracy assessment is a crucial step for validating the classification results of any supervised machine learning algorithms. Normally, accuracy assessment is carried out by comparing the results generated by a machine learning algorithm with the ground truth which is usually available. However, in the case of satellite images, we have to generate the ground truth first. To do that, we selected random 100 points for each class and labeled those points using the QGIS feature. Supervised learning was then applied to the images to perform the classification. Again 100 points were picked randomly from the classified data and that was visually assessed if the label had been correctly assigned or not. It's worthwhile to state here the overall accuracy of 94% was obtained for all the classes. The accuracy assessment and corresponding user's and producer's accuracies are presented in the Table 3 and Table 4.

| Reference Data | Water | Trees | Crops | Built Area | Bare Ground | Rangeland | Reference Total |
|---|---|---|---|---|---|---|---|
| Water | 98 | 0 | 0 | 2 | 0 | 0 | 100 |
| Trees | 2 | 96 | 2 | 0 | 0 | 0 | 100 |
| Crops | 0 | 1 | 97 | 1 | 1 | 0 | 100 |
| Built Area | 0 | 0 | 0 | 96 | 2 | 2 | 100 |
| Bare Ground | 0 | 0 | 0 | 4 | 96 | 0 | 100 |
| Rangeland | 1 | 0 | 0 | 2 | 1 | 96 | 100 |
| **Classified Total** | 101 | 97 | 99 | 105 | 100 | 98 | 600 |

**Table 3.** Classified Data Table with reference and classified totals.

| LULC Class | Reference Total | Classified Total | Number Correct | Producer Accuracy | User's Accuracy |
|---|---|---|---|---|---|
| Water | 100 | 101 | 98 | 98 | 97 |
| Trees | 100 | 97 | 96 | 96 | 99 |
| Crops | 100 | 99 | 97 | 97 | 98 |
| Built Area | 100 | 105 | 96 | 96 | 91 |
| Bare Ground | 100 | 100 | 96 | 96 | 96 |
| Rangeland | 100 | 98 | 96 | 96 | 98 |

**Table 4.** Accuracy Assessment for LULC Classification.

## 3. Results and Analysis

This section briefly explains the findings in terms of changes in built-up area and crop regions as discussed below.

*3.1. Changes in Built-up Areas*

As depicted in the figures 3 and 4 below, the built-up area has increased in all governorates of the country. The maximum increase has been observed in the North Batinah governorates while Wusta has observed the maximum increase in terms of percentage increase. It is worthwhile to note that the Muscat governorate has not seen any significant increase. This could possibly be due to the reason that the region is already mostly built-up and the mostly the vertical construction would have taken place in the specified period. Likewise, the maximum relative expansion in the Wusta governorate could be due to its small built-up baseline area.

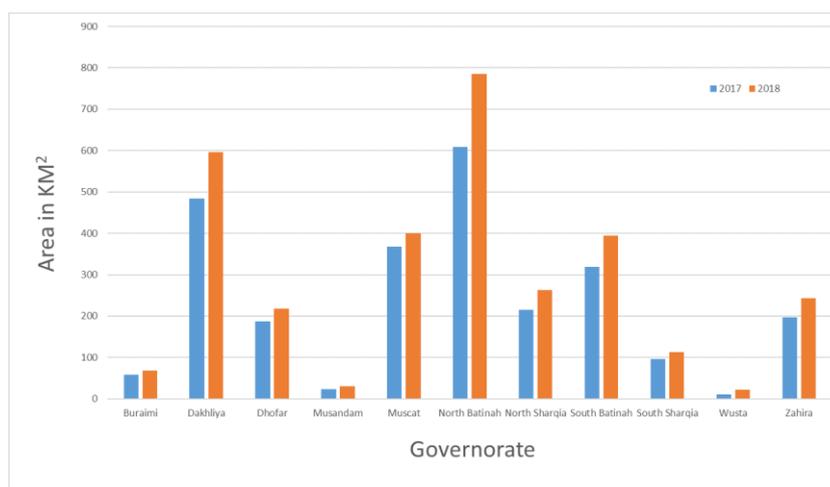

**Figure 3.** Built-up Area.

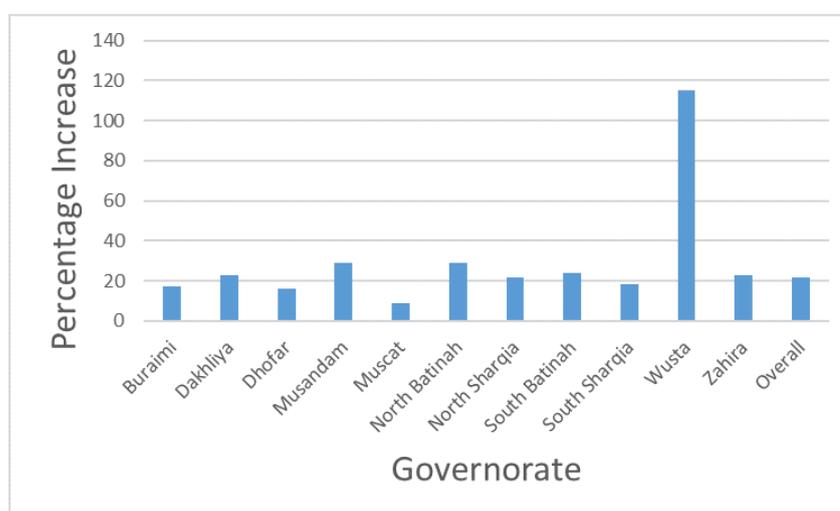

**Figure 4.** Percentage Increase in Built-up Area.

*3.2. Changes in Crop Area*

As shown in the Figure 5 and 6, the crop area has decreased in most of the governorates. The governorates of Dhofar and North Batinah have seen the maximum decrease in the crop area while some governorates have observed a marginal increase in the crop area. It is of utmost importance for the government to take steps so that the crop area and yield can be increased in the future. It is interesting to see the built-up area and the population are increasing at a speed much higher than that of the spread of crop area. This could be a point of concern for the authorities and hence there is a dire need to increase the crop area and hence increase the crop production.

## 4. Conclusion

This work utilized the publicly available Sentinel 2 satellite spatiotemporal data to analyze and compare the land cover land use changes across different governorates of Oman. After obtaining the data of the chosen area, a manual labelling process was carried out to create the ground truth data, followed by training of a supervised classification algorithm. An accuracy assessment was carried out to confirm the effectiveness of the trained model. All the data was then fed into the trained model to carry out the land use land cover classification. Detailed analysis were made to visualize and compare the changes in land use land cover across different governorates of Oman. This work demonstrates how the spatiotemporal satellite data could be used to obtain useful insights for effective urban planning.

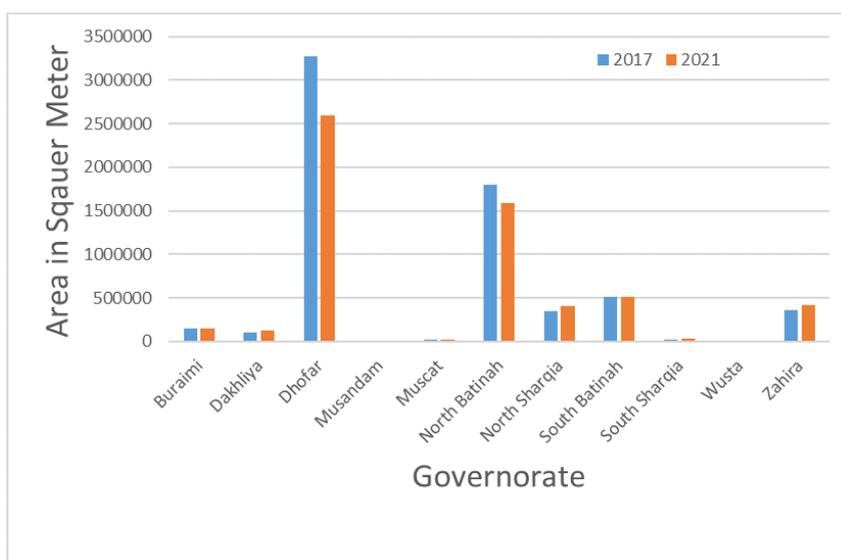

**Figure 5.** Crop Area.

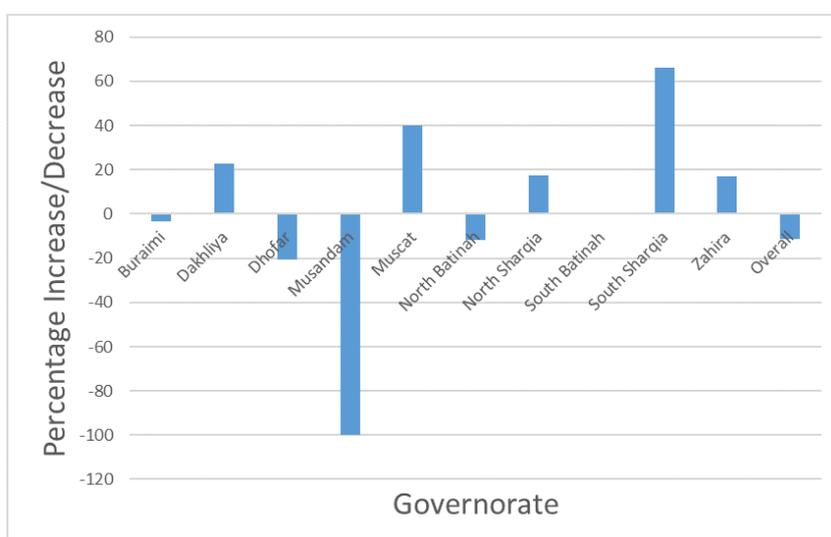

**Figure 6.** Percentage Change in Crop Area.